# Recent Advances in Graph-based Machine Learning for Applications in Smart Urban Transportation Systems

Hongde Wu[§], Sen Yan[§], Mingming Liu[†]

*School of Electronic Engineering and Insight SFI Centre for Data Analytics, Dublin City University, Dublin9, Ireland.*
[§] *Both authors contributed to the work equally and are joint first authors*
[†] *Corresponding author E-mail: mingming.liu@dcu.ie*

**Abstract:**
The Intelligent Transportation System (ITS) is an important part of modern transportation infrastructure, employing a combination of communication technology, information processing and control systems to manage transportation networks. This integration of various components such as roads, vehicles, and communication systems, is expected to improve efficiency and safety by providing better information, services, and coordination of transportation modes. In recent years, graph-based machine learning has become an increasingly important research focus in the field of ITS aiming at the development of complex, data-driven solutions to address various ITS-related challenges. This chapter presents background information on the key technical challenges for ITS design, along with a review of research methods ranging from classic statistical approaches to modern machine learning and deep learning-based approaches. Specifically, we provide an in-depth review of graph-based machine learning methods, including basic concepts of graphs, graph data representation, graph neural network architectures and their relation to ITS applications. Additionally, two case studies of graph-based ITS applications proposed in our recent work are presented in detail to demonstrate the potential of graph-based machine learning in the ITS domain.

**Abbreviations**

| | |
|---|---|
| ADAS | Advanced Driver Assistance Systems |
| AKF | Adaptive Kalman Filter |
| ARIMA | AutoRegressive Integrated Moving Average |
| AST-GCN | Attention-based Spatial-Temporal Graph Convolutional Network |
| CNN | Convolutional Neural Networks |
| ConvLSTM | Convolutional Long Short-Term Memory |
| DCRNN | Diffusion Convolutional Recurrent Neural Network |
| DL | Deep Learning |
| FC-LSTM | Fully Connected Long Short-Term Memory |
| GAT | Graph Attention Network |
| **GCN** | **Graph Convolutional Network** |
| **GNN** | **Graph Neural Network** |
| **ITS** | **Intelligent Transportation Systems** |
| **LSTM** | **Long Short-Term Memory** |
| **ML** | **Machine Learning** |
| **MLE** | **Maximum Likelihood Estimation** |
| **PSO** | **Particle Swarm Optimization** |
| **RF** | **Random Forest** |
| **STDN** | **Spatial-Temporal Dynamic Network** |
| **STGCN** | **Spatial-Temporal Graph Convolutional Network** |
| **ST-ResNet** | **Spatial-Temporal Residual Network** |
| **SVM** | **Support Vector Machine** |

## 1. Introduction

In recent years, due to the acceleration of urbanization, many people are moving to cities rapidly. In many countries around the world, especially developing ones, the growing demand for public transport services and the growing number of private vehicles are putting enormous pressure on existing transport systems. Frequent traffic accidents, serious traffic congestion, longer commuting time and other problems greatly reduce the efficiency of urban operations and affect the travel experience of passengers. To address these challenges, an increasing number of cities in the world have been developing intelligent transportation systems (ITS) to facilitate efficient traffic management by optimizing the utilization of system resources. Typically, an intelligent transportation system can leverage a set of technologies and tools including connected sensors, real-time communication, advanced control and optimization methods to allow a variety of road users to efficiently share information in the road networks. For instance, a roadside camera can be facilitated in a highway network to timely track moving vehicles by using computer vision techniques. Such a system can be used to accurately estimate the average speed of vehicles, predict future traffic flow as well as timely incident and congestion detection among many others.

At the heart of ITS applications is often driven by some learning-based algorithms and mechanisms which allow optimal decisions to be made for some traffic scenarios of interest. Specifically, graph-based machine learning methods have attracted tremendous interest and research efforts in recent years by researchers and practitioners worldwide thanks to their capability to effectively capture traffic data contained in the graph data structure. Currently, these algorithms have been increasingly applied to address various intractable challenges which were hardly tackled in the past. Therefore, our key objective of this chapter is to briefly review the algorithm designs in this field with a particular focus on graph-based learning approaches, including graph neural network (GNN) architectures. We will also present two recent applications leveraging GNNs to address challenges in traffic management with reference to our recent works (Chen et al., 2021c; Wu et al., 2022) before we conclude the chapter.

To begin with the chapter, we shall briefly introduce some background information related to this research subject by presenting some important research problems in the context of ITS and corresponding research directions with reference to the prior works in Veres and Moussa (2020); Jiang and Luo (2022).

**1.1 Traffic congestion and event detection**

Traffic congestion refers to the situation where the volume of traffic exceeds the design capacity of a road network (Nair et. al., 2019). This situation can often happen in large cities during peak time which will inevitably reduce the speeds of moving vehicles, leading to reduced traffic flow, longer travel time for commuters, as well as an increased level of pollution in the living environment (Chen et al., 2016; Zhang et al., 2011). To deal with such a challenge, traffic event detection aims to timely identify such a situation from road networks through various methods and tools such as road-side computer vision-based cameras, wireless sensor devices in vehicles and social media channels (Wang et al., 2017; Gu et al., 2016). The collected information can not only be used to provide real-time information to drivers to allow them to make decisions on alternative routes but also can be used to instruct traffic infrastructures, e.g., by coordinating traffic lights in a more effective manner, thereby reducing the number of vehicles on busy roads, and smoothing the traffic flows.

**1.2 Travel accidents and safety**

Traffic accidents are one of the reasons that endanger traffic safety. To improve traffic safety, it is not only necessary to improve people's safety awareness and enforce traffic policies against dangerous driving behaviors, but more importantly, to devise effective technology and means to help prevent accidents from happening at the early stage. Along this line, the development of advanced driver assistance systems (ADAS) has been largely adopted and integrated into modern vehicles, including but not limited to speed advisory systems (Chen et al., 2021a; Liu et al., 2016), parking assist (Liu et al., 2019a), lane changing detection (Waykole et al., 2021; Wang et al., 2020) and adaptive cruise control (Hajek et al., 2013; Vahidi and Eskandarian, 2003). Recent research directions have been focused on pedestrian crossing prediction (Cadena et al., 2022), autonomous vehicle collision prediction (Malawade et al., 2022), and vehicle trajectory prediction (Jo et al., 2021).

**1.3 Traffic demand prediction and resource relocation**

Travel demand refers to the amount of travel that travelers make when choosing transportation services to meet their travelling requests. From the perspective of traffic management agencies, it is important to gather such information for better planning and scheduling of traffic resources. For instance, frequent trains should be scheduled during peak time between commuting towns and more bus routes should be facilitated to areas where travel demand is high. Travel demand prediction focuses on the problems of knowing future travel behavior and pattern of travelers according to historical data and current trends such as weather and events. Often, such problems can be formulated as time series forecasting problems (Torres et al., 2021) where spatial and temporal aspects can be taken into consideration along with other socioeconomic factors, e.g., place of interest. The prediction problems have recently attracted great attention in the research community with the rapid growth of sharing economy (Chen et al., 2021b). On the one hand, an accurate travel demand forecasting model can help to share companies better manage their assets in this ever-growing competitive market. On the other hand, an accurate travel demand prediction model can help enhance the quality of experience for customers which in turn can help increase profits for the sharing companies. Finally, we note that a destination/route prediction problem is related to the travel demand prediction problem, but they are distinguishable. The travel destination prediction problem aims to estimate the number of travelers that will visit a specific location at a particular time, while the destination/route prediction problem aims to address the prediction challenge of the future destination/route for a given traveler. We refer interested readers to the following works in this research direction (Krumm et al., 2013; Epperlein et al., 2018; Taguchi et al., 2018; Liu et al., 2019b; Yan et al., 2022).

**1.4 Automatic and autonomous driving**

One of the key design aspects of ITS is to increase the level of autonomy for vehicles driving on the roads. Generally speaking, this can be done depending on if a human driver is involved in steering the vehicle and to what level the driver needs to operate the vehicle under specific conditions. More specifically, automatic driving refers to the use of technology, including dedicated software, sensors, and algorithms to assist a human driver in operating a vehicle. In this setup, a driver is still in control of the vehicle while some functions may be taken over by an automatic unit. For example, a cruising control system can automatically control the moving speed of a vehicle once a desirable cruising speed is set by the driver (Shaout and Jarrah, 1997). Autonomous driving, on the other hand, refers to self-driving vehicles which can operate without human input or intervention. Clearly, this requires a higher level of autonomy and more advanced control for the vehicle to make the optimal decision at the right time and space, thus requiring more design efforts for various engineering challenges. In the context of ITS, graph-based visual recognition algorithms have been applied in some popular research directions, such as object detection for vehicles (Tayara et al., 2018), lanes and pedestrians (Li et al., 2015); trajectory prediction for vehicle driving (Li et al., 2020). It is expected that with the increasing level of connectivity and integration in ITS in the coming years, self-driving vehicles will play an important role in transportation systems by improving traffic safety, efficiency as well as sustainability (Urmson and Whittaker, 2008; Bagloee et al., 2016; Goldbach et al., 2022).

The rest of the chapter is organized as follows. We first review some existing methods including statistical methods and machine learning methods which have been widely used to address the challenges as aforementioned. After that, we shall introduce graph-based machine learning and demonstrate their efficacy in dealing with the challenges with a particular focus on graph neural networks (GNNs). Two recent applications of GNNs will also be presented to detail the benefits of GNNs in real-world ITS scenarios before we conclude the chapter.

**2. Research Methods in Transportation Data Analysis**

In this section, we briefly review existing research methods in the transportation domain. Roughly speaking, current methods reported in the literature can be summarized into the following three categories, namely, statistical methods, machine learning (ML) methods and deep learning (DL) methods.

Statistical methods refer to the use of some well-established mathematical techniques and principles in statistics to analyze, interpret and make inferences about a population based on a sample or subset of data (Longnecker and Lyman, 2015). Among a large set of statistical methods, Kalman filtering (Okutani and Stephanedes, 1984; Wang and Papageorgiou, 2005; Guo et al., 2014), maximum likelihood estimation, i.e., MLE (Spiess, 1987; Watling, 1994; Xie et al., 2007; Jamal et al., 2022), and AutoRegressive Integrated Moving Average, i.e., ARIMA (Ye et al., 2019), are the most popular and typical options for studies in transportation data. Specifically, based on real-world traffic flow data, Guo et al. (2014) proposed and applied Adaptive Kalman Filters (AKFs), which can update the process variances on the prediction of the stochastic short-term traffic flow rate. The results then demonstrated improved adaptability of AKFs compared to traditional Kalman filters, especially for highly volatile traffic data. On the other hand, Jamal et al. (2022) adopted an MLE method in the estimation of vehicle critical gaps under heterogeneous traffic conditions. Tested on the dataset collected in Peshawar, Pakistan, the MLE method's results were proven to assist in distinguishing between cars, and two- and three-wheeled vehicles.

ML methods, on the other hand, aim to learn patterns contained in data and make predictions for unseen data using the identified patterns without means of explicit programming. ML models can better model complex nonlinear relationships contained in traffic data which may not be easily done using traditional statistical methods (Boukerche et al., 2020; Vlahogianni and Karlaftis, 2013). To illustrate a few, ML algorithms including Support Vector Machine, i.e., SVM (Wang and Shi, 2013; Fu et al., 2016; Ling et al., 2017; Tang et al., 2019), K-Means (Montazeri-Gh and Fotouhi, 2011; Kadir et al., 2018) and Random Forest, i.e., RF (Ou et al., 2017) have been applied in the transportation data studies. For instance, Ling et al. (2017) combined multi-kernel SVM and adaptive Particle Swarm Optimisation (PSO) to predict short-term traffic flow and the results demonstrated the feasibility and superiority of the proposed scheme compared with four baseline models which were based on other different PSO algorithms. Additionally, with the aim of traffic condition recognition, Montazeri-Gh and Fotouhi (2011) applied the K-means clustering algorithm to identify the driving features of hybrid electric vehicle drivers and cluster the driving segments accordingly. Adopting appropriate features, K-means achieved satisfactory results with an accuracy rate of 87%.

DL methods, a subfield of ML, can also learn patterns from data in the transportation domain. However, compared to traditional ML models, such as RF and SVM, DL models can capture more complex relationships from traffic big data through neural network architectures using multiple hidden layers and high-level representation of data which may not be easily designed by humans (Gadri et al., 2021). Existing DL algorithms, such as Convolutional Neural Networks, i.e., CNNs (Dabiri and Heaslip, 2018; Chen et al., 2018; Nguyen et al., 2020), and Long Short-Term Memory, i.e., LSTM (Tian and Pan, 2015; Zhang et al., 2018; Nawaz et al., 2020), have been widely considered for sequence modelling and time series forecasting tasks thanks to their powerful capacity in capturing correlation in data. In this context, Chen et al., (2018) first proposed a multiple 3D CNNs architecture (MST3D) based on spatiotemporal correlation for citywide vehicle flow prediction. Being able to capture the correlation of spatial and multiple temporal dependencies, the proposed method performed better than the state-of-the-art baselines, such as Spatio-Temporal Residual Networks (ST-ResNet) and Spatial-Temporal Dynamic Networks (STDN). Furthermore, Nawaz et al. (2020) proposed a ConvLSTM model that blended the concepts of both CNN and LSTM into a single model and applied it in transportation mode learning to extract spatiotemporal features from trajectory data and weather conditions. The comparative studies revealed the superiority of the proposed ConvLSTM architecture where at least 3% in accuracy had been improved over other benchmark models involving SVM, RF and CNN.

As a concluding remark for this section, we note that our review presented in this section is not exhaustive and hereby we refer interested readers to some related survey papers and comprehensive research works for each subfield of research methods to appreciate the big picture of these studies. To name a few, for the statistical methods, Faouzi et al. (2011) reviewed data fusion techniques adopted in multiple tasks in transportation systems such as automatic incident detection and traffic demand estimation, Molugaram and Shanker Rao (2017) introduced a full range of data analysis topics and provided worked-out examples and solved problems for a wide variety of transportation engineering challenges, and Washington et al. (2020) described and illustrated several statistical and econometric tools commonly used in transportation data analysis. Regarding ML and DL techniques, Alsrehin et. al. (2019) explored and reviewed the data mining and ML technologies in transportation issues, Zantalis et al. (2019); Li and Xu (2020); Yuan et al. (2021) focused on ML techniques in intelligent transportation systems, Hillel et al. (2021) reviewed ML classification methodologies for modelling passenger mode choice, Tsolaki et al. (2022) study various ML algorithms for freight transportation and logistics applications, Ahmed and Diaz (2022) discussed ML methods in urban mobility and Huang et al. (2023) explored and summarized context-aware ML models applied in ITS. Additionally, Veres and Moussa (2020); Haghighat et al (2020); Haydari and Yilmaz (2022) mainly focused on DL methods and applications in smart transportation systems.

## 3. Graph Neural Networks for Transportation Data Modelling

Graph neural networks (GNNs) are a special type of deep neural networks which aim to process data contained in graph-based structures. They have recently achieved great success in many research areas including but not limited to physical system modelling, machine translation, object detection as well as cloud application (Sanchez-Gonzalez et al., 2018; Kipf et al., 2018; Beck et al., 2018; Hu et al., 2018; Nguyen et al., 2022), thanks to their capability in processing data in non-Euclidean spaces. Similarly, graph-powered ML models have been applied in the transportation domain to help address challenges such as traffic state prediction (Zheng et al., 2020; Wu and Liu, 2022), travel demand prediction (Zhuang et al., 2022) and traffic anomaly detection (Wu et al., 2022). Recent research works have shown that GNN-based architectures can achieve better performance than most DL methods (Lin et al., 2020; Bui et al., 2023). To provide a clear understanding of the mechanism of GNNs, we review the basic concepts of GNNs in this section with reference to our previous work (Nguyen et al., 2022).

### 3.1 Graph data representation

A graph $G$ is typically represented by a set of vertices $V$ (i.e., nodes) and a set of edges $E$ (i.e., links). Mathematically, the graph is defined by $G = (V, E)$. For a pair of vertices $u, v \in V$, we use $u \rightarrow v$ to denote a directed edge $e_{uv} = (u, v) \in E$ between $u$ and $v$ which forms an ordered pair of nodes $(u, v) \in V \times V$ in a directed graph. Specifically, we define the neighbour relationship as follows. In a graph $G = (V, E)$, for a pair of vertices $u$ and $v$ in $V$, $u$ could be called a neighbour of $v$ if and only if there exists an edge $e_{uv} = (u,v)$ in $E$. Additionally, we define $N(v)$ as the set of all neighbours of $v$ in $G$. Mathematically, $\forall u, v \in V, \exists e_{uv} = (u,v) \in E \Leftrightarrow u \in N(v)$. The connections between all nodes in $G$ can be represented by an adjacency matrix $A^{n \times n}$ with $A_{ij} = 1$ if $e_{ij} \in E$ or $A_{ij} = 0$ otherwise, where $n = |V|$ denotes the number of nodes.

In the context of ITS, the adjacency matrix A can be used to indicate whether a pair of nodes in the graph data is connected. Although physical constraints may fix the connectivity of nodes (i.e., a road intersection is connected to the neighbor road intersection by a road segment physically), designers can usually define the value of the adjacency matrix, and even dynamically train the adjacency matrix from the dataset. However, in addition to physical constraints, it is important to understand how to use an adjacency matrix to best capture the connectivity between different nodes in the graph.

Roughly speaking, four types of adjacency matrices have been studied in the literature, namely space ($S$), time ($T$), space-time ($ST$) and adaptive ($A$). The spatial adjacency matrix is usually based on distance. Euclidean distance (Yu et al., 2018; Chen et al., 2020) or physical geographic distance (Kim et al., 2019) between different sites (i.e., nodes in the graph) is usually used to calculate the weight of their entries. A temporal adjacency matrix can be defined based on the similarity, i.e., Pearson correlation coefficient (Bai et al., 2019) between the temporal data of each pair of nodes/stations (i.e., historical traffic demand sequence). To merge the benefits of both spatial and temporal features, spatial-temporal embedding can be generated for each node in the graph (Ye et al., 2020). However, in this case, it is difficult to intuitively describe the adjacency matrix with high-dimensional embedded features, therefore the adaptive adjacency matrix has been proposed (Wu et al., 2019), where the value of the adjacency matrix can be adaptively defined through the training process of GNNs.

### 3.2 Graph neural networks

GNNs are deep neural networks that are suitable for analyzing graph-structured data. In this section, we review the prevalent variants of GNNs for ITS, including Graph convolutional networks (GCNs), Spatial temporal graph convolutional networks (STGCN) as well as Graph attention networks (GAT).

### 3.2.1 Graph convolutional networks (GCNs)

Spectrum theory is a branch of mathematics study that can be applied to graphs. It represents graphs by a normalized Laplace matrix. Through a series of calculations of the Laplace matrix, namely graph convolution operation, the information of the graph will be extracted (Bruna et al., 2013). In GCNs, the value of a filter, which is with a set of learnable parameters, can be determined through the training process of GCNs. Also, GCNs can be seen as a space-based method. Specifically, GCN can be considered as the aggregation of feature information from the neighboring nodes. Some recent mathematical studies have improved GCNs by exploring the alternative representation of the Laplace matrix.

### 3.2.2 Spatial-temporal graph convolutional networks (STGCN)

STGCNs target the hidden properties of the spatial-temporal graphs such that both spatial dependency and temporal dependency can be captured from the data in graphs simultaneously. This characteristic is particularly useful to address practical challenges, such as traffic-flow prediction, as data generated in a traffic network can be easily modelled using a spatial-temporal graph. In this context, the STGCN (Yu et al., 2018) is composed of several spatial-temporal convolutional blocks, each of which is formed with two gated sequential convolution layers and one spatial graph convolution layer in between. Spatial-temporal convolutional blocks were introduced and applied repeatedly in this architecture, combining several graph convolutional layers with sequential convolution in order to represent the spatial-temporal relations.

### 3.2.3 Graph attention networks (GAT)

Attention is a popular technique in DL that mimics physiological cognitive attention. The effect enhances the importance of small parts of the input data and de-emphasis the rest. This technique has been used to enhance the prediction performance for many sequence-based tasks of GNNs, i.e., GAT (Veličković et al., 2018). For instance, a temporal attention mechanism can add an importance score for each historical time step to measure the influence and this strategy can effectively improve the accuracy of prediction accuracy (Chen et al., 2021c).

### 3.2.4 Graph-based learning

Given the representation of graphs we introduced above, graph-based learning tasks could be divided into three levels: node level, edge level, and graph level. Firstly, at the node level, by employing a multi-perceptron or a SoftMax layer as the output layer, GNNs are used to solve classification tasks (i.e., sorting nodes into respective classes), prediction tasks (i.e., predicting the continuous values on nodes), and clustering tasks (i.e., grouping similar nodes together). Secondly, in edge-level tasks, GNNs are usually used in the determination of the edge connection strength, the prediction of the link between two nodes, and the classification of edges. Lastly, graph-level tasks involve graph classification, graph regression, and graph matching, with GNN layers combined with pooling layers and/or readout layers to produce an abstract representation at the graph level.

## 3.3 GNNs for Emerging ITS Applications

In this section, we will detail two emerging ITS applications including bike availability prediction (Chen et al., 2021b) and lane detection (Wu and Liu, 2022) based on our recent research to showcase the efficacy of GNNs in addressing practical challenges arising in ITS. In each case study, we present our research problem and objective and demonstrate how GNN can be applied to address the problem of interest.

### 3.4 Bike availability prediction

There is a growing interest in adopting bike-sharing systems globally. However, the traditional bike management methods can easily lead to supply-demand imbalance due to estimation errors of system operators and unexpected traffic delays during the bike transition. Thus, due to the uncertainty of departure and arrival of bikes at any bike station, it is important to take a more proactive approach by accurately predicting the number of bikes that will be available for users to access at any given time and location. For instance, in Figure 1, each node is a bike station and each blue number in the circle indicates the number of available bikes in real time. In this scenario, we wish to predict the number of available bikes at each station in a near future time slot (e.g., in 15 minutes).

This problem can be formulated as follows. Given a bike-sharing system consisting of $N$ bike stations, $N \coloneqq \{1,2,\ldots,N\}$ indicates a set of station indexes. $B_t^i \in R$ indicates the number of available bikes at the station $i$ at the time $t$. Let $B_t \in R$ be the vector of available bikes across all $N$ stations at the time $t$. Let $F_t \in R^{N \times d}$ be the feature space for all $N$ stations at the time $t$, where the number of features associated with each station is $d$. Our objective is to find the learning function $H(\cdot)$ which can address the following problem:

$$B_{t+1:t+n} = H(B_{t-m+1:t}; F_{t-m+1:t})$$

where $m$ and $n$ denotes the input and output length for the model respectively and $t+1:t+n$ presents the sequence output from time step $t+1$ to $t+n$. We note that this problem is formulated as a sequence-to-sequence prediction problem as the input is featured by a sequence of past observations of available bikes. The model outputs the predicted number of available bikes in a sequential manner, capturing the future n slots of available bikes at any given station. Here, the graph-based learning function $H(\cdot)$ is taken as a GAT (Chen et al., 2021c), and both spatial and temporal aspects of the data have been captured in the work. The underlying graph is based on an undirected graph: $G = (V, E, A)$, where $V$ is the set of nodes (i.e., bike stations), $E$ is the set of edges, indicating the connection between stations, and $A$ is the adjacency matrix, which is used to model the relation between different stations, e.g., distance-based metrics. The proposed attention-based STGCN (AST-GCN) model has been applied to predict the number of available bikes using two real-world datasets. Compared to baseline models, such as XGBoost, FC-LSTM (fully connected LSTM) and DCRNN (Diffusion Convolutional Recurrent Neural Network), AST-GCN can achieve performance with a mean absolute error equals 1 in the best setup for the dataset based in Dublin city.

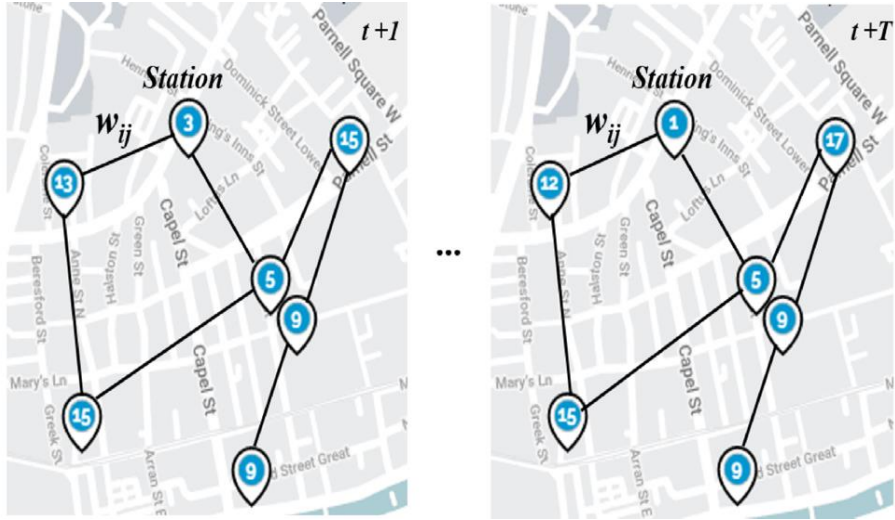

Figure 1: A subset of bike stations of the Dublin bike-sharing system operates in real-time. The edges are added for the illustration of the inherent graph signals. (*Reprinted from "A Comparative Study of Using Spatial-Temporal Graph Convolutional Networks for Predicting Availability in Bike Sharing Schemes", by Chen, Z. et al., (2021).*)

### 3.5 Lane change intention prediction

In transportation, predicting the lane-change intention can help prevent potential threats to driving safety. In our case study, we select a specified road segment in the highway road networks shown in Figure 2 and we assume that vehicles can move freely with frequent acceleration, deceleration, and lane-changing behavior. Our objective is to predict the probability of lane change for drivers. To this end, we monitor the average driving speed and the number of vehicles on each lane per second using a sliding window-based method for data collection. More information on the research background can be referred to the work of Wu and Liu, (2022).

Here, the collected traffic flow data are modelled using graphs. More specifically, we use $G = (V, E)$ to represent a highway network, where $V$ denotes the nodes representing the set of lane segments $V = \{l_i | i = 1,2, \dots, N\}$, where $N$ denotes the maximum number of nodes in the graph. Let $E$ be edges representing connections between a pair of nodes in graph $G$. The adjacency matrix is denoted by $A$. The connectivity of the graph is set as fully connected as the vehicle may change lanes from one to any other while driving on the road segment without traffic surveillance. To better illustrate this point, Figure 2 demonstrates our detailed modelling process. The highway network is divided into two road segments with each segment consisting of four lanes, i.e., eight lanes in total $N = 8$. The feature set for lane $i$ at each time step $t$ is denoted by $X_i(t) = (S_i(t), D_i(t))$, where $D_i(t)$ and $S_i(t)$ denote the number of vehicles and the average moving speed on the lane at time, respectively. Finally, the length of the input time window for model training is denoted by $T$ for each lane segment. With this setting, we collect the historical values (i.e., the data from previous time duration) of the number of vehicles and the average moving speed on each lane so that the probability of lane changes can be predicted in the near future, i.e., the next 90 seconds. In comparison with the baseline model, temporal-CNN, the proposed prediction scheme, Lane-GNN, has shown its superior capability in detecting drivers' lane-changing intention which can achieve an accuracy of 99.42% within 90 seconds in the specific experimental settings.

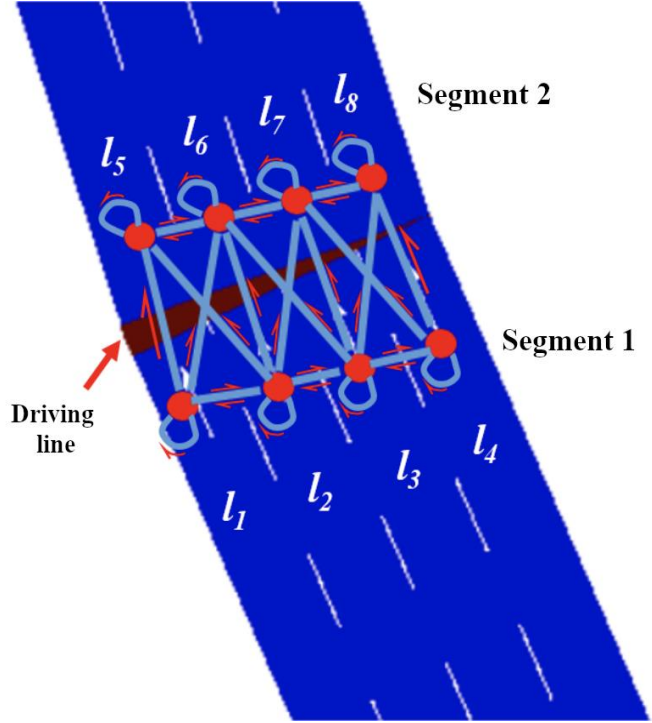

Figure 2: Graph modelling of two consecutive road segments of the $M50$ highway network. Each lane denotes a node (red points) and the connectivity between different lanes denotes an edge (blue lines). A vehicle can change lanes or remain in its current lane depending on its driving intention. (*Reprinted from "Lane-GNN: Integrating GNN for Predicting Drivers' Lane Change Intention", by Wu, H. and Liu, M. (2022).*)

## 4. Conclusion

Graph-based machine learning has made advances in addressing modern traffic-related challenges. This is an emerging research area and has recently attracted many efforts in the research community. To motivate our discussion, in this chapter, we first presented the background information on several key technical challenges for the design of ITS. After that, we briefly reviewed the research methods in the literature that have been applied to address these challenges, starting from classic statistical methods to modern machine learning and deep learning-based approaches. Following that, our research focus has been primarily on graph-based machine learning methods. This was highlighted through a detailed review of basic concepts of graphs, graph data representation, graph neural network architectures as well as their links to ITS applications. Finally, we presented two GNN-based ITS applications based on our recent works and demonstrated their efficacy in dealing with the challenges in the real world by comparing their performance with other baseline models.

## Acknowledgements

The publication has emanated from research supported in part by Science Foundation Ireland under Grant Number *21/FFP-P/10266* and *SFI/12/RC/2289_P2*, co-funded by the European Regional Development Fund.

## Author Biography

**Hongde Wu** received his B.Eng degree in Biomedical Engineering with an outstanding graduate from Beijing Institute of Technology in 2017 and received his M.Eng degree (research degree), under the supervision of Dr. Mingming Liu, from Dublin City University in 2022. Currently he works as a researcher in 2012 Lab at Huawei. His interests are deep learning, graph neural network and smart transportation.

**Sen Yan** is a Ph.D. student at the School of Electronic Engineering and Insight SFI Centre for Data Analytics, Dublin City University, under the supervision of Dr. Mingming Liu and Prof. Noel E. O'Connor. He received the B.Eng in Internet of Things Engineering from Shandong University, China in 2018 and M.Sc in Computer Science from Trinity College Dublin in 2020. Sen is working in the areas of intelligent transportation and micromobility. His research interest also includes image processing, computer vision, and machine learning.

**Mingming Liu** is an Assistant Professor in the School of Electronic Engineering at Dublin City University (DCU). He is also affiliated with the SFI Insight Centre for Data Analytics as a Funded Investigator. He received the B. Eng (1st Hons) in Electronic Engineering from National University of Ireland Maynooth in 2011 and the PhD in Control Engineering from the Hamilton Institute at the same university in 2015 when he was 26. He is an IEEE senior member and has published over 50 papers to date, including "IEEE Transactions on Smart Grids", "IEEE Transactions on Intelligent Transportation Systems", "IEEE Transactions on Automation Science and Engineering", "IEEE System Journal", "IEEE Transactions on Transportation Electrification", "Scientific Reports" and "Automatica". Since 2018, he has secured more than 1.2 million euros in research funding as the sole PI. Currently, he is the management committee member in Ireland for EU COST Actions CA19126, CA20138 and CA21131. His research interests include control, optimization and machine learning with applications to 5G & IoT, electric vehicles, smart grids, smart transportation, smart healthcare and smart cities.